\documentclass[preprint,12pt,fleqn]{elsarticle}

\usepackage{hyperref}
\usepackage[utf8]{inputenc}
\usepackage{times}
\usepackage{graphicx}
\usepackage{float}
\usepackage{algorithm2e}
\usepackage{array}
\usepackage{tabularx}
\usepackage{listings}
\usepackage{booktabs}
\usepackage{subcaption}

\usepackage{amsmath}
\usepackage{amsfonts}
\usepackage{amssymb}

\newcommand{\sentence}[1]{"\emph{#1}"}
\newcommand{\annotation}[1]{\textbf{\footnotesize{#1}}}
\newcommand{\annot}[2]{(\emph{#1}\slash\annotation{#2})}

\newcommand{\cad}{\noexpand\hbox{c'est-à-dire }}

\journal{Swarm and Evolutionary Computation}

\begin{document}

\begin{frontmatter}

\title{Comparison of Global Algorithms in Word Sense Disambiguation}
\author{Loïc Vial, Andon Tchechmedjiev, Didier Schwab \\ 
LIG-GETALP\\
Univ. Grenoble Alpes\\
$\lbrace$Loic.Vial, Andon.Tchechmedjiev, Didier.Schwab$\rbrace$@imag.fr \\
\url{http://getalp.imag.fr/xwiki/bin/view/WSD/}
}

\begin{abstract}
This article compares four probabilistic algorithms (global algorithms) for Word Sense Disambiguation (WSD) in terms of the number of scorer calls (local algorithm) and the F1 score as determined by a gold-standard scorer. Two algorithms come from the state of the art, a Simulated Annealing Algorithm (SAA) and a Genetic Algorithm (GA) as well as two algorithms that we first adapt from WSD that are state of the art probabilistic search algorithms, namely a Cuckoo search algorithm (CSA) and a Bat Search algorithm (BS). 
As WSD requires to evaluate exponentially many word sense combinations (with branching factors of up to 6 or more), probabilistic algorithms allow to find approximate solution in a tractable time by sampling the search space. We find that CSA, GA and SA all eventually converge to similar results (0.98 F1 score), but CSA gets there faster (in fewer scorer calls) and reaches up to 0.95 F1 before SA in fewer scorer calls. In BA a strict convergence criterion prevents it from reaching above 0.89 F1. 
\end{abstract}


\end{frontmatter}

\section{Introduction}
Word Sense Disambiguation (WSD) is a complex task that consists in finding the best sense of each word of a text with relation to the surrounding words (context). It is a fundamental problem in Natural Language Processing, which can help in performing tasks like the automatic extraction of multilingual information or machine translation. 
A lot of research has already been carried out on WSD. Consequently, there are many approaches, including fully supervised methods (make use of sense-annotated corpora to train supervised classifiers) and similarity-based methods (rely on dictionaries, thesauri and more generally knowledge sources). Fully supervised methods require large hand-annotated corpora, a somewhat rare and expensive resource that must be crafted specifically for a particular sense inventory, language even domain. Moreover, a supervised classifier requires at least 1000 examples/word to achieve a stable performance, while there are at most 200/word in the largest sense-annotated corpora \cite{Miller:1993:SC:1075671.1075742}. Due to this bottleneck, similarity-based methods perform on par with supervised algorithms on standard evaluation datasets. While there are few prospects for improvement with supervised system, similarity-based systems offer many avenues for improvement even for English (the best resourced language) \citep{vial-el-al-2016}.

In this article, we focus on similarity-based methods, which rely on the notion of global and local algorithms. Local algorithms compute a similarity measure between two senses of two words of the text. Global algorithms make use of the local similarities to find the most suitable sense for each word of the text. The execution time is exponential in the size of the input, thus requiring the use of a window of limited width around each word when evaluating sense combination. The problem can become intractable even for short sentences: a linguistically motivated context, such as a paragraph for instance would be intractable. Probabilistic approaches attempt to alleviate the intractability by exploring only a sampling of the search space in order to find an approximate solution as close to the optimal solution as possible. Probabilistic sampling approaches have been successfully used for decades for complex optimisation where the geometry (and the precise formula) of the search function are unknown (non differentiable). Two classical algorithms that belong to this category are Simulated Annealing and Genetic Algorithms and have been successfully used for WSD in past years. However, since the appearance of SA and GA there have been many new and more efficient probabilistic optimisation algorithms, in particular two of them that are at the state of the art of the field (in classical computational model, thus excluding quantum versions).   However, Bat Optimisation and Cuckoo Optimisation have never been applied to Word Sense Disambiguation, a combinatorial problem more difficult in magnitudes (number of parameters and branching factor of the search space) compared to classical NP-complete combinatorial problem (e.g. the travelling salesman algorithm).

After a presentation of similarity-based methods for Word Sense Disambiguation, we present existing global algorithms and in particular the Simulated Annealing Algorithm (SAA) and the Genetic Algorithm. We then introduce our adaptation of Bat Search and Cuckoo Optimisation for WSD.

\section{Similarity-based Word Sense Disambiguation}

Similarity-based word sense disambiguation rests on the notion of \emph{local algorithm} and \emph{global algorithm} introduced by Schwab et al. \cite{schwab2}.

A local algorithm gives a similarity measure between two pairs of \annot{word}{sense}. For example, considering the sentence \sentence{Your algorithm contains many bugs}, if we compute the similarity measure between the pair \annot{algorithm}{piece of code} and \annot{bugs}{insect} using a local algorithm, we expect to have a smaller value than if we compute the similarity measure between the pair \annot{algorithm}{piece of code} and \annot{bugs}{fault in a system}.

Then, a global algorithm will propagate these local similarity measures to an upper level, in order to achieve the actual disambiguation of a complete document.
In general, global algorithms are heuristic. Indeed, in a large document, we cannot compute the local similarity measure algorithm for every possible \annot{word}{sense} pair.  For example,  if we consider a 500 word text with 5 senses for each word, there are $500^5$ possibilities. Even if the local algorithm could run in 1 nanosecond (which it clearly doesn't), it would still take approximately one year to evaluate all combinations

Global algorithms represent a particular combination of sense assignments as a \emph{configuration} and make use of a \emph{scoring function} (a.k.a. scorer or fitness function) to evaluate the fitness of the configuration.

A configuration is the set of all \annot{word}{sense} associations, for every word in a document, we say that a configuration annotates each word of the text with a sense. That is, for example, if a document contains 100 words that we need to annotate with senses, a possible configuration for this document is a vector of dimension 100, where for every index $i \in [0,100[$  (the position of the word in the document) is assigned an integer $j$ in the range of the possible senses indexes for this word.

A configuration scorer, scorer, or fitness function, is a measurement of the quality of a configuration. Generally, it can be as simple as the sum of all similarity measure for every \annot{word}{sense} pair in a configuration, using a particular local algorithm.

Given these definitions, and knowing that we cannot compute the fitness score of every single combination in the search space, global algorithms use heuristics to maximise the exploration of the configuration space while minimizing the number of evaluations of the fitness function.

This paper will focus on the compared performance of four different global algorithms: a Simulated Annealing Algorithm (SAA), a Genetic Algorithm (GA), a Bat Algorithm (BA) and a Cuckoo Search Algorithm (CSA).

\section{Existing Probabilistic Global Algorithms for WSD}
Many experiments that compare heuristic global algorithms for disambiguation have been carried out. In particular, we can take note of the work of Schwab et al. \cite{schwab1,schwab2}, where a SAA and a GA (neighbourhood/local search approaches) were implemented for the WSD problem, with the objective of comparing them to an Ant Colony Algorithm (a constructive approach).
We will reuse the existing implementations of SAA and GA in order to compare them to the two new algorithms that we have adapted for WSD. SAA and GA are briefly described in the following. All data and programs are accessible through the companion page of this article\footnote{\url{http://getalp.imag.fr/static/wsd/Vial-et-al-global-howto/}}.

\subsection{The Simulated Annealing Algorithm}

Simulated annealing for Word Sense Disambiguation has been introduced by Cowie et al. \cite{Cowie1992}. This approach is inspired by the annealing process in metallurgy, where the sturdiness of the metal is increased by controlling the rate of cooling applied to it \cite{laarhoven1987simulated}.
In context of WSD, the ``metal'' is simply the configuration of all \annot{word}{sense} pairs in the text.

The algorithm works in cycles. Each cycle is composed of iterations. At each iteration a single random change is made in the configuration: the sense assigned to a random word in the text is changed to another sense randomly (uniform distribution) and the score of the modified configuration is computed. If the new configuration has a higher score than the configuration before the modification, it replaces the previous configuration. Otherwise, we compute a probability of accepting the new configuration anyway despite the fact it has a lower score. The acceptance probability is classically computed following a Boltzmann distribution. Simulated annealing is a particular instance of the Metropolis-Hastings algorithm \citep{Metropolis1953}. 
The strategy of accepting lower scores is a mechanism that allows escaping from local minima that typically abound in complex highly dimensional search spaces.

\subsection{The Genetic Algorithm}
Genetic algorithms are evolutionary algorithm based on the principle of the natural selection of species, and more particularly ``Survival of the Fittest''.\\
The idea of GA is to consider search space configurations as individuals in a population. 

GA starts with an initial population, where a certain number of the fittest individuals (as evaluated by the fitness function) will be crossed with each other (crossover operation), that is parts of the configurations will be swapped between two individuals. 

Then, a certain number of random changes (mutation operator]) are applied to the population (like for SAA but repeated). Finally a subset of the fittest the new population are selected and kept (elitism). Variants of GA copy the best individual in proportion to their fitness in the new population. A convergence criterion is used to determine when the population converges and when there are no more beneficial evolutions taking place. 

GA was first adapted to WSD by Gelbukh et al.\cite{Gelbukh2003} where the population is a set of configurations, and the fitness function is the configuration scorer.

\section{Two New Global Algorithms for WSD}
Among of non-quantum probabilistic search algorithms, Bat Search and Cu\-ckoo Search offer the best overall performance (speed with relation to approximation quality), both invented by Xin-She Yang. As these algorithms have never been adapted to WSD and given that in many applications they are better than other similar algorithm, we have decided to propose an adaptation to solve the WSD problem. 
We will describe them, their advantages, and our implementation as global algorithms for WSD in the following.

\subsection{The Bat Algorithm}

The Bat Algorithm was originally introduced by Yang \cite{bat1} and is inspired by they way bats move through echolocation in order to locate and capture pray. These bats emit a very loud ultrasound pulse (inaudible to humans) and listen for the echo that bounces back on the boundaries of the environment and of the objects therein in order to pinpoint their position and the positions of their prey. A sound pulse is characterised by its frequency, rate of emission and loudness. There are different sound pulses depending on the situation and context. When the bat is searching for prey, the loudness is at its highest and the rate of emission is very low (rough sketch of the environment). Once a prey is located, the bat lowers the loudness and increase the rate of emission in order to track the prey with precision.  

The particularity of this algorithm is its ability to converge to a promising region of the search space rapidly while concurrently exploring alternative areas.

\subsubsection{Description of the Algorithm}

The behaviour is highly configurable through many parameters. 

The minimum and maximum frequency $f_{min}, f_{max}$ of the bats sound pulse define the scale of the movement speed of the bats. Large values are more suitable for large space. 

The minimum and maximum $\lambda_{min}, \lambda{max}$ loudness of the sound pulses have an impact on the width of the local search.

$\alpha$ and $\gamma$ are threshold parameters to influence when the bat decides to converge to a local solution when it is  \emph{close enough} to its "prey" (When its score is better than the current best score).

Let the current iteration be $i$. Let the set of all bats be $B$ and a bat $b\in B$. Let $b_{current} \in B$ be the current bat in the iteration and let $b_{best} \in B$ be the bat with the best fitness.

Let $score: B \mapsto \mathbb{R}$ be a function that evaluates the fitness of a bat. Let the frequency be denoted by $f$.
Let $r(min,max): \mathbb{R}^2 \mapsto [min, max] \in \mathbb{R}$ be a function that yields a random integer drawn from the uniform distribution with a value between min and max. 

Let the velocity be noted $\nu$ and let $p: B \mapsto \mathbb{R}$ be a function that gives the relative position of a bat $b$ expressed by a real number. Let the loudness be $\lambda$ and the average loudness across $B$ be $\bar{\lambda}$. Let the pulse rate be noted $\rho$.

\RestyleAlgo{boxruled}
\begin{algorithm}[tbh!]
\SetAlgoLined
Create a population of bats\;
For each bat is assigned a random position, a null velocity, a random pulse frequency, a random pulse loudness and a random pulse rate\;
\While{stop condition is not met}{
Rank the bats and find the best one according to the objective function\;
\ForEach{bat}{
Update the pulse frequency, the position and the velocity of the bat:
\[f= r(f_{min},f_{max})\]
\[\nu = \nu + (p(b_{best}) - p(b_{current}))\cdot f\]
\[p(b_{current}) = p(b_{current}) + \nu\]

\If{$r(-1,1) > \rho$}{
Fly randomly around the best bat
\[p(b_{current}) = p(b_{best}) + r(-1,1) \cdot \bar{\lambda}\]
}
\If{$r(-1,1)<\rho$ \textbf{and} $score(b) > score(b_{best})$}{
Accept the new solution\;
Increase the bat pulse rate and reduce the bat pulse loudness
\[\lambda = \lambda \cdot \alpha\]
\[\rho = \rho \cdot (1 - e^{-\gamma * i})\]
}
}
}
Return the best bat\;
\caption{Bat Algorithm pseudocode}
\label{alg:bat}
\end{algorithm}%
\autoref{alg:bat} summarizes the Bat Search algorithm in pseudo code.

\subsection{The Cuckoo Search Algorithm}
\subsubsection{General idea}
The Cuckoo Search Algorithm is another creation of Yang \cite{cuckoo1}. The idea comes from the aggressive reproduction strategy of some cuckoo birds: they lay their eggs in the nests of other host birds (of other species). If the host bird later discovers that an foreign egg was laid its nest, it will either discard it, or just abandon its nest to build another elsewhere.

\subsubsection{Description of the Algorithm}
Cuckoo search is described in pseudo-code in \autoref{alg:cuckoo}.

\begin{algorithm}[tb]
\SetAlgoLined
Generate a population of $n$ cuckoos\;
\While{convergence criterion not met}{
Draw a cuckoo $i$ among $n$ randomly (uniform)\;
Move the cuckoo pseudo-randomly following a Lévy flight\;
Choose another cuckoo $j$ among $n$ randomly (uniform)\;
Replace $j$ by $i$ if $score(i) > score(j)$\;
Sort the cuckoos by their score\;
Replace the cuckoos with the worst scores by new randomly generated cuckoos\;
}
Return the cuckoo with the best score\;
\caption{Cuckoo Search Algorithm pseudocode}
\label{alg:cuckoo}
\end{algorithm}

\subsubsection{The Importance of Lévy Flights}
A fundamental aspect of Cuckoo search is the use of the Lévy flights to determine the movements of the cuckoos (sample the search space). A Lévy flight is a random walk following a Lévy distribution. The Lévy distribution is an interesting probability distribution. The Lévy distribution yields values most of the time in proximity to zero (when not shifted) but sometimes a significant larger value.
The $c$ parameter is the Lévy scale, the only parameter of the distribution. There is an optional shift called Lévy location, that centres the distribution around 0.
%
%
The Lévy flight is important because it allows the algorithm to intensively explore local regions of the search space and to sometimes jump in far regions that allows to escape local minima.
\subsection{Adaptation to WSD}
In order to tailor implementations of those algorithms as global algorithms for WSD, we need to make a number of assumptions and modifications to the original algorithms.

The original algorithms specify the movement of bats or cuckoos with a vector with real coordinated (characterize by a direction and a norm). Given that WSD is a discrete problem, the definitions of what a movement means must be adapted as a vector with real coordinates does not exist. Moreover all senses are semantically distinct and independent and there isn't any meaningful relative distance between them, thus moving from sense \#1 to sense \#2 is no less significant than moving from sense \#1 to sense \#4.  For example, in the WordNet 3.1 lexical database, the sense \#1 of the noun "mouse" corresponds to the animal, whereas the sense \#2, \#3 and \#4 correspond respectively to "a black eye", "a person who is timid" and "a computer device". It is clear that the sense \#2 is not closer to the sense \#1 than the sense \#3 or \#4.
 The position of the bats in the BA and of the cuckoos in the CSA in the search space is represented by the configuration that assigns a sense to each word of the text \annot{word}{sense} for a given text (that is, a configuration is a particular solution to our problem).
We adapt the original definition by using the velocity of bats and the Lévy flight for cuckoos to change the current position of the agents by making random changes to their correspond configuration just like for GA and SA.

For example, a flight of distance 5 will have the effect of performing 5 random changes in the configuration of a cuckoo. The same is true for the velocity of bats.

The algorithms were implemented in Java\footnote{\url{http://getalp.imag.fr/static/wsd/Vial-et-al-global-howto/}} and a detailed code is provided in \autoref{appendix:implem}.

\section{Comparing Global Algorithms}
\subsection{Method}

We compared the four algorithms using the evaluation corpus from the coarse-grained English all-words WSD task from SemEval-2007  campaign \citep{semeval2007Task7}.\\
The corpus is composed of 5 sense annotated texts from different domains and sources:
\begin{itemize}
\item The first, second and third texts come from the Wall Street Journal and are respectively 368, 379 and 500 words long.
\item The fourth text is the entirety of the Wikipedia entry for Computer Programming and is 677 words long.
\item The fifth text is an excerpt from the novel "Knights of the Art" by Amy Steedman and is 345 words long.
\end{itemize}
The campaign contain a reference annotation that we use to calculate how good a particular configuration is in terms of the F1 score. The two criteria we use for our evaluation is the F1 score of the results with relation to the number of invocation of the scoring function as we want to obtain the best solutions in the least amount of time.

\subsubsection{The Choice of the Configuration Scorer}
As we explained in the introduction, a global algorithm is a method to explore effectively the space of possible configurations, without trying every single possibility of pair \annot{word}{sense} in a document.\\
To do so, a global algorithm needs a configuration scorer which helps by evaluating the quality of a configuration by giving a score value for a configuration, in order to see if a configuration is better or worse than another one in terms of the objective.

A classical objective function is to use a semantic similarity measure and to sum all pairwise similarities between selected senses in a configuration. For example, in previous implementation of SAA and GA for WSD, the Lesk similarity measure that calculates the overlap between the definitions of senses \citep{lesk} was used.  Given that we are merely interested in the global algorithms, in this article we make the choice to use the F1 score from the gold standard scorer as the objective function for the optimisation, in other words, the local algorithm is an oracle from the point of view of the local algorithm so as to reduce the problem from AI-complete to an NP-complete optimisation problem and to be able to evaluate only the influence of the global algorithm.

\subsubsection{Parameters Tuning}
Both BA and CSA have numerous parameters that must be tuned for each specific problem and have a significant influence on the outcome.\\
In our implementation of the algorithms, we have the following parameters:
\begin{itemize}
\item The BA has 7 parameters: the number of bats, the minimum frequency of their sound pulses, the maximum frequency of their sound pulses, the minimum loudness of their sound pulses, the maximum loudness of their sound pulses, the smoothing coefficients $\alpha$ and $\gamma$. 
\item The CSA has 4 parameters: the number of cuckoos, the number cuckoos destroyed per iteration, the location parameter of the Lévy distribution and the scale parameter of the Lévy distribution.
\item The GA has 3 parameters: the size of the population, the crossover rate and the mutation rate.
\item The SAA has 2 parameters: the cooling rate, number of iterations, the initial acceptance probability (the initial temperature is estimated to match the initial acceptance probability). We fixed the initial acceptance probability to 0.8.
\end{itemize}

In order to find the optimal parameters for each algorithm without manually testing every combination, we implemented a Cuckoo Search Algorithm that automatically finds the optimal parameters.

We run the algorithms many times (e.g. 100--1000) in order to obtain a representative sample of the distribution of possible answers, and use the Mann-Whitney-U statistical test  \cite{Mann47} in addition to the score similarly to the modified Simulated Annealing algorithm used for parameter estimation by \citet{tchechmedjievetal2012}.

We perform the parameter estimation on the first two of each document of the SemEval 2007 task 7 corpus, or a total of 94 words to annotate. In order to avoid any bias in the results, we remove the sentences used for parameter estimation from the evaluation corpus. Moreover, we perform the estimation several times with different limits on the number of scorer calls rather than a traditional convergence criterion: 200, 800, 2000, 4000, so that we can have an idea of the optimal parameters and the resulting F1 scores we could obtain with varying computational constraints.

\subsubsection{Evaluation}
After finding the optimal parameters for each algorithm, we evaluate their efficiency by comparing the best configuration score (F1 score) they can obtain in function of the number of calls to the scorer (200, 800, 2000 and 4000). For each algorithm and each scorer call threshold we run the algorithm 100 times and plot the average F1 score across the whole corpus and the 100 runs compared to the average number of evaluations of the scoring function. 

\subsection{Results and Discussion}
The results of optimal parameters estimation are presented in \autoref{tab:bestbat} for the Bat Algorithm, in \autoref{tab:bestcuckoo} for Cuckoo Search, in \autoref{tab:bestga} for the Genetic Algorithm and in \autoref{tab:bestsa} with several thresholds for the number of calls to the scorer as a convergence criterion.	

\begin{table}[H]
\centering
\begin{tabular}{@{} c c c c c c c c @{}}
\toprule
\textbf{\# calls to scorer} & \textbf{\# bats} & \textbf{mn. fq.} & \textbf{mx. fq.} & \textbf{mn. ld.} & \textbf{mx. ld.} & \textbf{$\alpha$} & \textbf{$\gamma$} \\
\midrule
200 & 50 & 45 & 100 & 23 & 24 & 0.40 & 0.14 \\
800 & 50 & 38 & 38 & 6 & 10 & 0.65 & 0.95 \\
2000 & 50 & 28 & 100 & 0 & 38 & 0.1 & 0.72 \\
4000 & 50 & 0 & 100 & 0 & 38 & 0.1 & 0.95 \\
\bottomrule
\end{tabular}
\caption{Optimal parameters for the Bat Algorithm}
\label{tab:bestbat}
\end{table}

\begin{table}[H]
\centering

\begin{tabular}{@{} c c c c c @{}}
\toprule
\textbf{\# calls to scorer} & \textbf{\# cuckoos} &  \textbf{\# cuckoos destroyed} & \textbf{Lévy location} & \textbf{Lévy scale} \\
\midrule
200 & 1 & 0 & 20 & 5 \\
800 & 1 & 0 & 0.37 & 5 \\
2000 & 1 & 0 & 5 & 0.5 \\
4000 & 1 & 0 & 5 & 0.5 \\
\bottomrule
\end{tabular}
\caption{Optimal parameters for the Cuckoo Optimisation Algorithm}
\label{tab:bestcuckoo}
\end{table}

\begin{table}[H]
\begin{center}
\begin{tabular}{@{} c c c c c @{}}
\toprule
\textbf{\# calls to scorer} & \textbf{population} & \textbf{crossover rate} & \textbf{mutation rate} \\
\midrule
200 & 100 & 0.02 & 0.01 \\
800 & 100 & 0.01 & 0.01 \\
2000 & 100 & 0.01 & 0.01 \\
4000 & 73 & 0.01 & 0.01 \\
\bottomrule
\end{tabular}
\end{center}
\caption{Optimal parameters for the Genetic Algorithm}
\label{tab:bestga}
\end{table}

\begin{table}[H]
\begin{center}
\begin{tabular}{@{} c c c @{}}
\toprule
\textbf{\# calls to scorer} & \textbf{cooling rate} & \textbf{\# iterations} \\
\midrule
200 & 0.95 & 1 \\
800 & 0.1 & 100 \\
2000 & 0.1 & 50 \\
4000 & 0.1 & 77 \\
\bottomrule
\end{tabular}
\end{center}
\caption{Optimal parameters for the Simulated Annealing Algorithm}
\label{tab:bestsa}
\end{table}

\begin{figure}
\begin{subfigure}[b]{.5\textwidth}
\includegraphics[width=\textwidth]{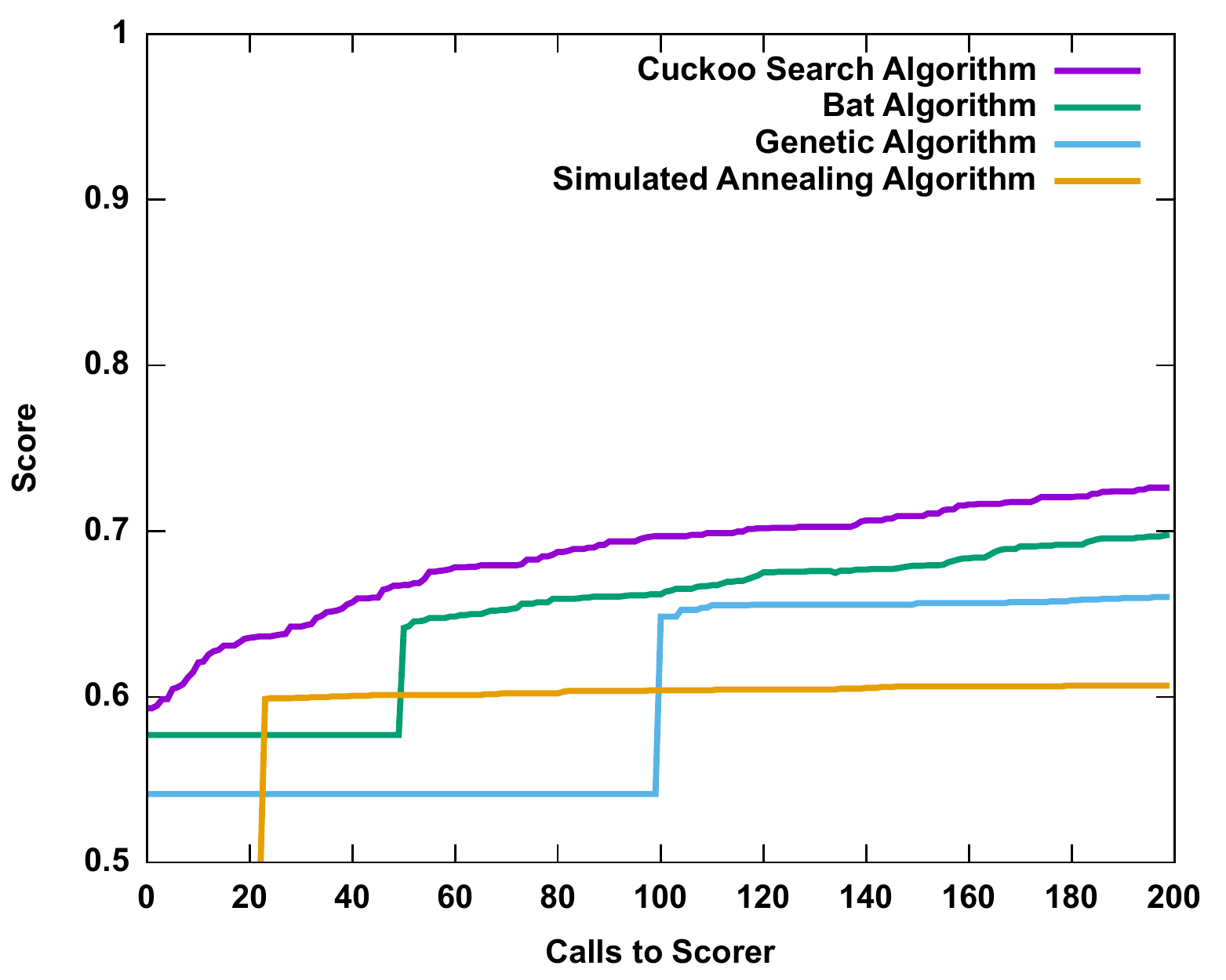}
\caption{200 scorer calls limit}
\label{lab:resultssum200}
\end{subfigure}
\begin{subfigure}[b]{.5\textwidth}
\includegraphics[width=\textwidth]{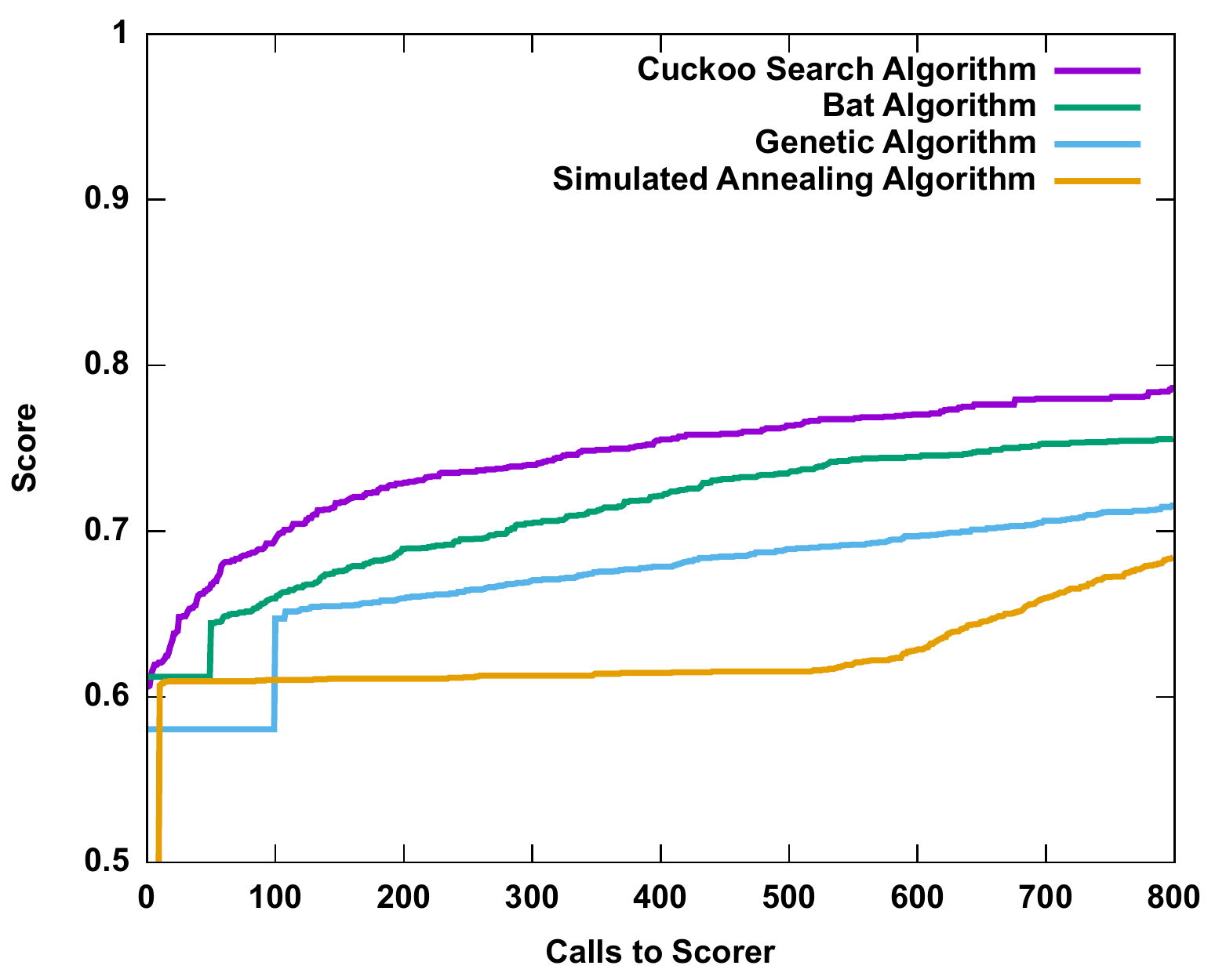}
\caption{800 scorer calls limit}
\label{lab:resultssum800}
\end{subfigure}
\begin{subfigure}[b]{.5\textwidth}
\includegraphics[width=\textwidth]{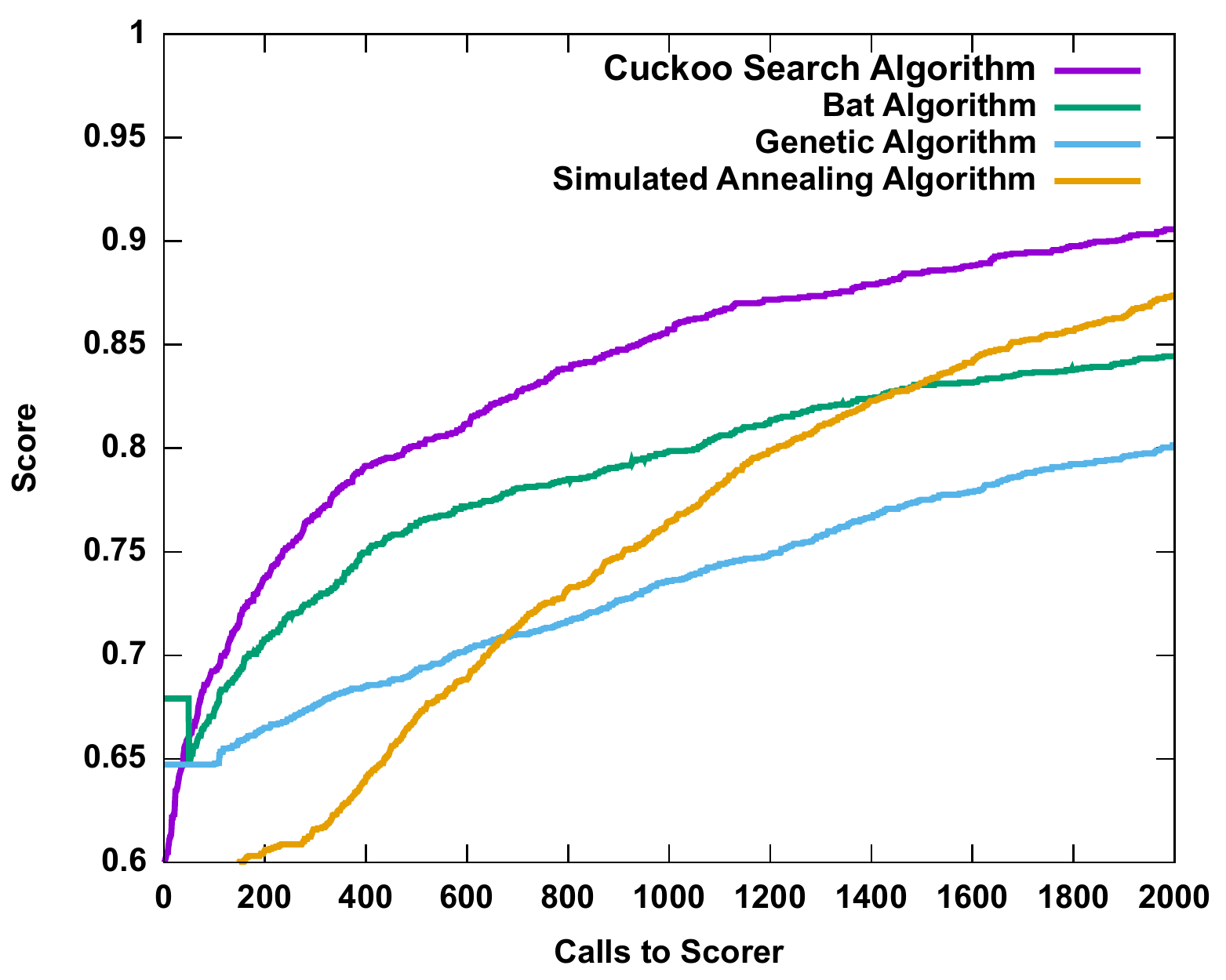}
\caption{2000 scorer calls limit}
\label{lab:resultssum2000}
\end{subfigure}
\begin{subfigure}[b]{.5\textwidth}
\includegraphics[width=\textwidth]{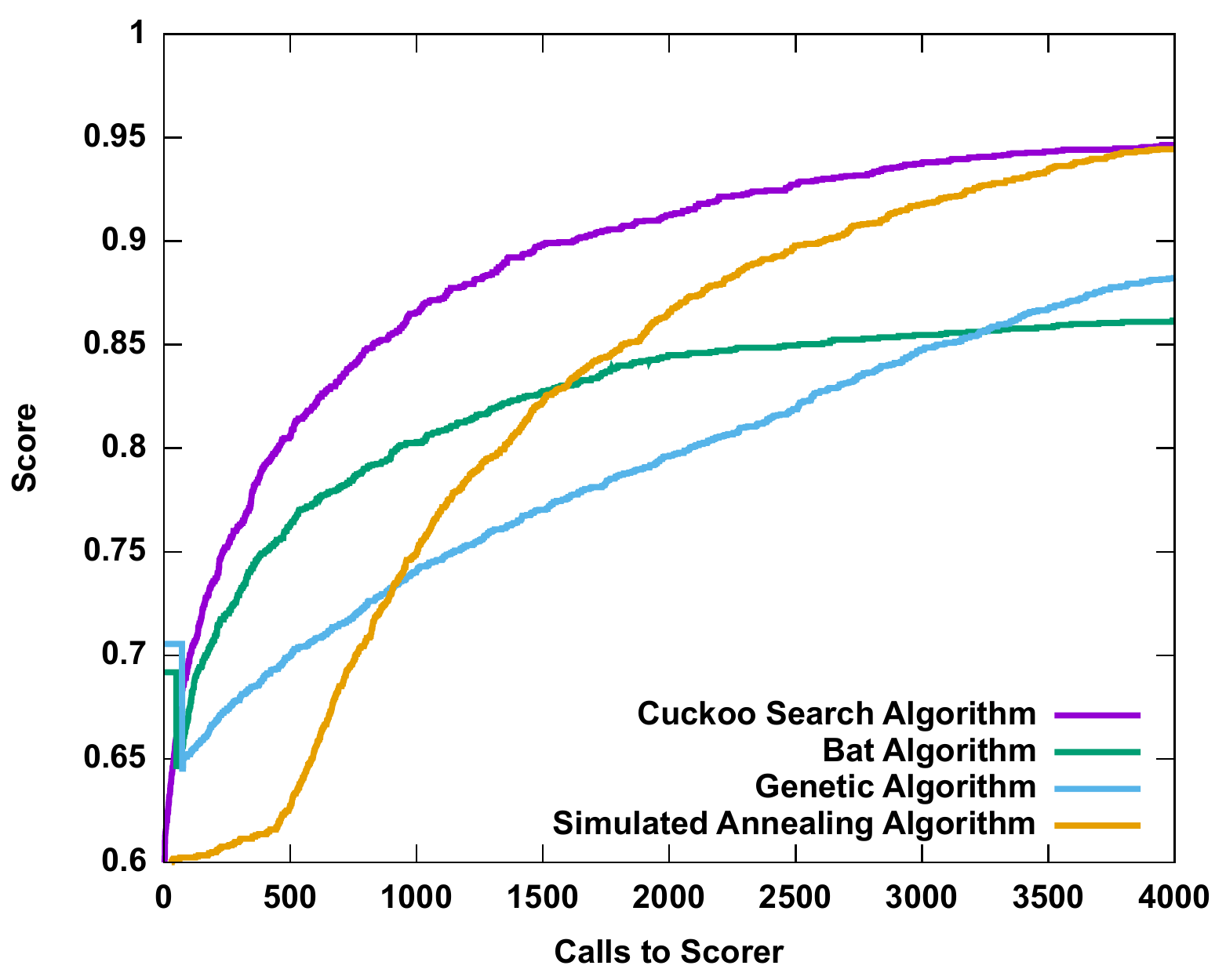}
\caption{4000 scorer calls limit}
\label{lab:resultssum4000}
\end{subfigure}
\begin{subfigure}[b]{.5\textwidth}
\includegraphics[width=\textwidth]{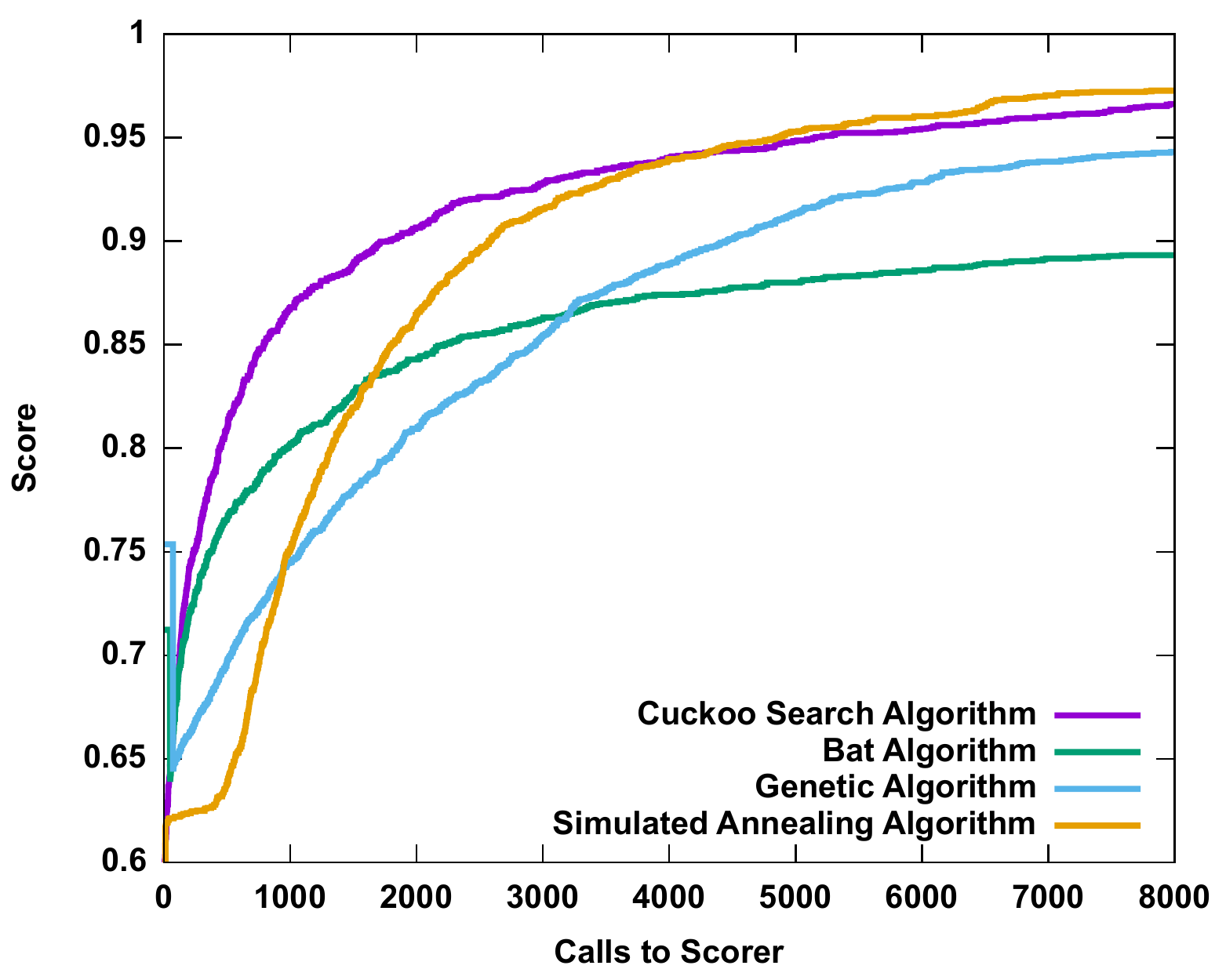}
\caption{8000 scorer calls limit}
\label{lab:resultssum8000}
\end{subfigure}
\begin{subfigure}[b]{.5\textwidth}
\includegraphics[width=\textwidth]{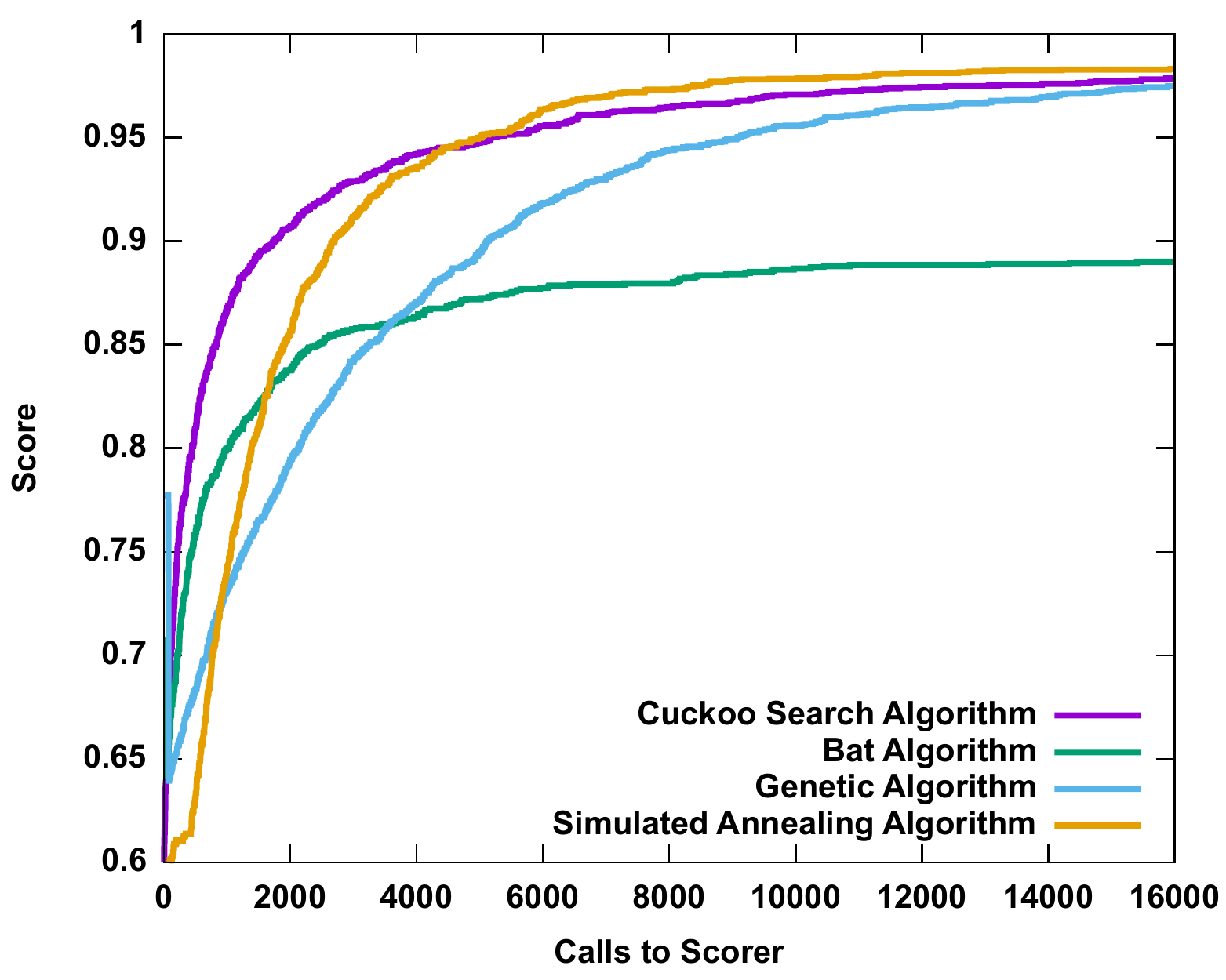}
\caption{16000 scorer calls limit}
\label{lab:resultssum16000}
\end{subfigure}
\caption{Overall average results for all the documents of the evaluation campaign corpus with 200, 800, 2000, 4000, 8000 and 16000 scorer calls}
\label{fig:resultssum}
\end{figure}

\autoref{fig:resultssum} presents the average overall results for all for algorithms for 200 (\autoref{lab:resultssum200}), 800(\autoref{lab:resultssum800}), 2000 (\autoref{lab:resultssum2000}) and 4000 (\autoref{lab:resultssum4000}) scorer calls. Given that with 4000 calls, the appearance of the charts suggest that convergence has not occurred. Thus, we make two additional executions with a limit of 8000 (\autoref{lab:resultssum8000}) and 16000 (\autoref{lab:resultssum16000}) scorer calls, with the same parameters as the run with the limit at 4000 scorer calls. 

We see that across the 6 experiments, the progression of the average F1 score compared to the average number of scorer calls remains the same, thus we will make a sweeping analysis across all size experiments. Lower limits to the number of scorer calls can be seen as a different "zoom levels" at the beginning of the execution.

Below 20 calls, SA cannot yet obtain a solution, while CSA obtains the best F1 (0.6)  followed by BS (0.59) followed by GA (0.55). After 100 scorer calls, the order stabilizes and we have CSA first (0.7), followed by BS (0.68) followed by GA (0.65) followed by SA (0.6). Until 500 calls, all algorithms except SA follow a slow increase (from 0.70 to 0.79 for CSA, from 0.68 to 0.71 for BA, from 0.65 to 0.67 for GA) while SA stagnates at 0.60. After 500 the algorithm continue the trend, however SA catches up rapidly and starts to overtake GA at 600 scorer calls (0.68 F1) and BS at 1400 scorer calls (0.8 F1 and stays above them). GA overtakes bat search after 3400 scorer calls and SA seems to overtake CSA after 4000 scorer evaluations. 

In order to better see if SA continues to rise after overtaking CSA, we perform two additional experiments with limits of 8000 and then 16000 scorer calls. In the long run The bat algorithm converges slightly below 0.90 F1	while, CSA, GA and SA all converge (0.98 F1) and continue rising, likely reaching a F1 of 1 eventually.

While all algorithms save for BS all converge to the same solutions, for lower counts of scorer calls, CSA gets better results with fewer scorer evaluations. This will be particularly advantageous for scorer functions that are costly to compute and where simulated annealing or GA will be prohibitive. Indeed, for these experiments we used a perfect scoring function that is very fast to calculate. However, typical heuristic scoring functions used when there is no gold-standard data available, such as Lesk are much more costly and may not allow to reach enough scorer calls to reach the point where SA overtakes CSA is a tractable time. 

As for the lacklustre performance of BS, after a point (depending on the loudness parameter) BS stops accepting the configurations even if they are better (the target is acquired by the bat), which likely explains the early convergence of the algorithm.

\section{Conclusion}
In this paper, we adapt two state of the art probabilistic search algorithms as global algorithms the Word Sense Disambiguation problem : a Bat Algorithm and a Cuckoo Search Algorithm. We further adapt Cuckoo search to perform parameter estimation for probabilistic global WSD algorithms.
We compared the two algorithms to two existing implementation of classical probabilistic optimisation algorithms, a Simulated Annealing Algorithm and a Genetic Algorithm. We make a comparative evaluation using the F1 score computed from the gold standard of the Semeval 2007 Task 7 WSD task against the number of calls to the scoring function. We used an oracle objective function in order to only evaluate the influence of the global algorithm on the results, rather than letting the heuristic scoring function have an influence as well.
We find that CSA, SA and GA all converge at around 16000 scorer calls and that SA slightly overtakes CSA after 4000 calls, while CSA gets there faster (in much fewer scorer calls). While the scorer used here was perfect and fast to compute (comparison to the gold-standard), actual heuristic scorer are much slower and may not allow to tractably reach the point where SA overtakes CSA. Thus, we conclude that CSA is a much better sampler of the WSD search space and that it should be preferred to SA and GA. We also conclude that because BS stops accepting solutions after a while (inherent to the algorithm, not an explicit convergence criterion) the convergence is too fast. 

In this work we only considered a perfect scorer, however we currently work on evaluating and comparing with an actual heuristic scoring function and on studying the correlation between the oracle objective function and the heuristic scoring function. This study will then allow to optimize and adapt the heuristic scoring function to be more like the oracle scoring function on the same subset of the data we used to evaluate the parameters for the global algorithms.


\bibliography{biblio-Vial-et-al-global}
\bibliographystyle{elsarticle-harv}

\cleardoublepage

\appendix
\section{Detailed results of the parameter estimation}
\label{appendix:params}
\subsection{Number of call to the scorer : 200}
\begin{figure}[H]
\begin{center}
\begin{subfigure}[b]{0.38\textwidth}
\includegraphics[width=\textwidth]{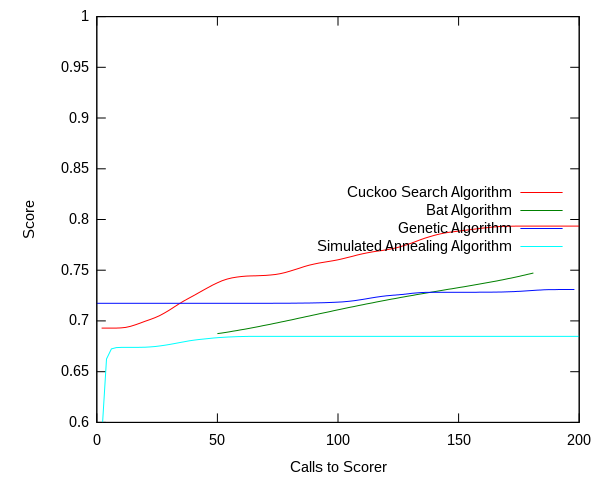}
\caption{Document 1}
\end{subfigure}
\begin{subfigure}[b]{0.38\textwidth}
\includegraphics[width=\textwidth]{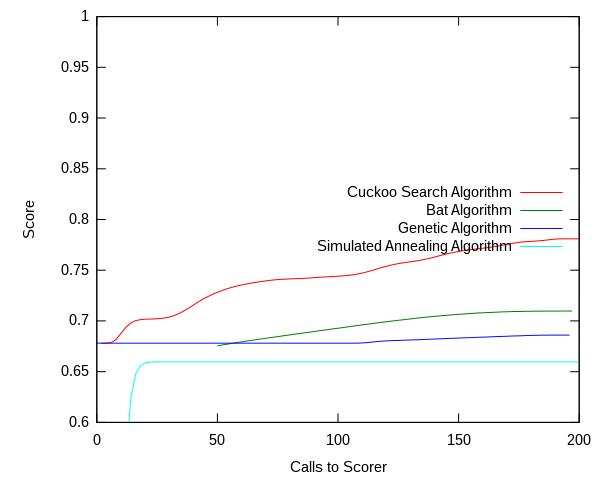}
\caption{Document 2}
\end{subfigure}
\begin{subfigure}[b]{0.38\textwidth}
\includegraphics[width=\textwidth]{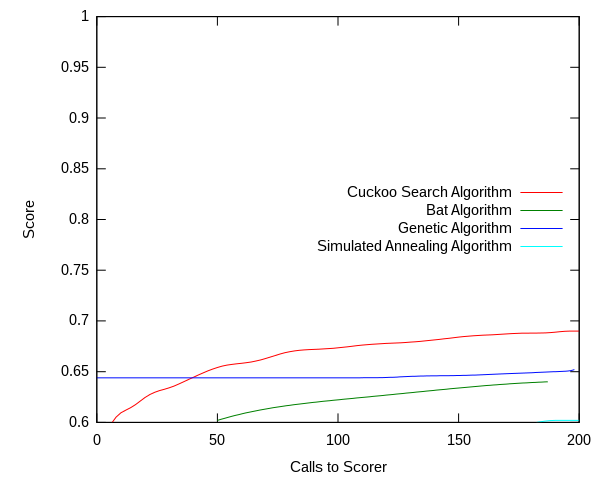}
\caption{Document 3}
\end{subfigure}
\begin{subfigure}[b]{0.38\textwidth}
\includegraphics[width=\textwidth]{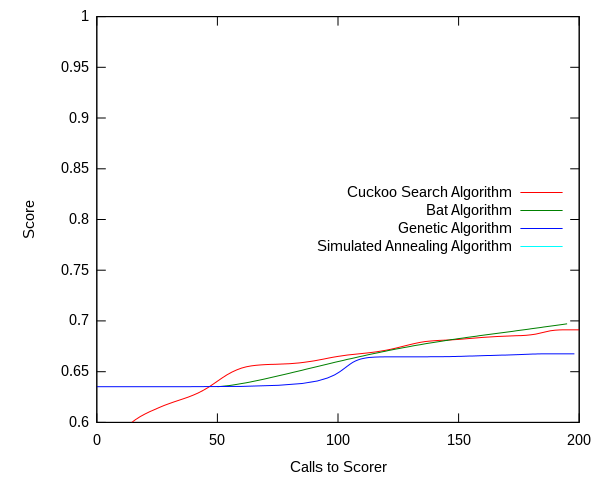}
\caption{Document 4}
\end{subfigure}
\begin{subfigure}[b]{0.38\textwidth}
\includegraphics[width=\textwidth]{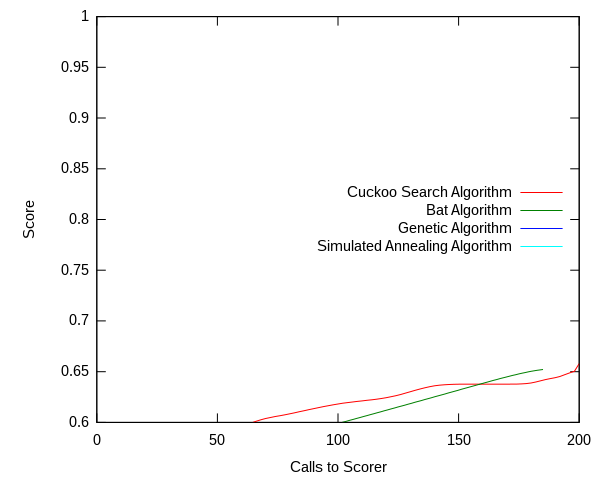}
\caption{Document 5}
\end{subfigure}
\end{center}
\caption{Evaluation of the global algorithms for 200 scorer calls}
\label{fig:ressco200}
\end{figure}

\newpage

\subsection{Number of call to the scorer : 800}
\begin{figure}[H]
\begin{center}
\begin{subfigure}[b]{0.38\textwidth}
\includegraphics[width=\textwidth]{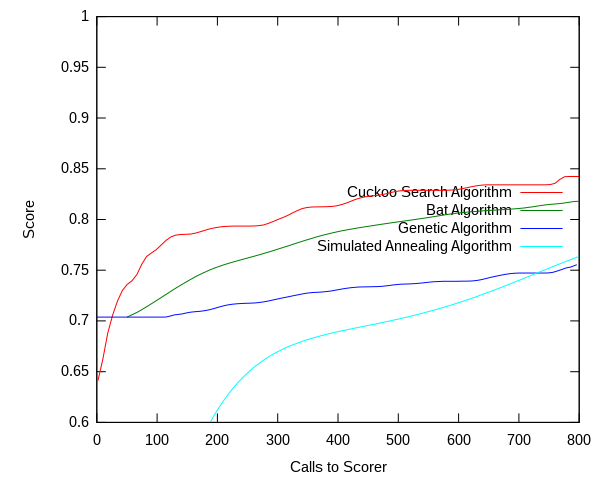}
\caption{Document 1}
\end{subfigure}
\begin{subfigure}[b]{0.38\textwidth}
\includegraphics[width=\textwidth]{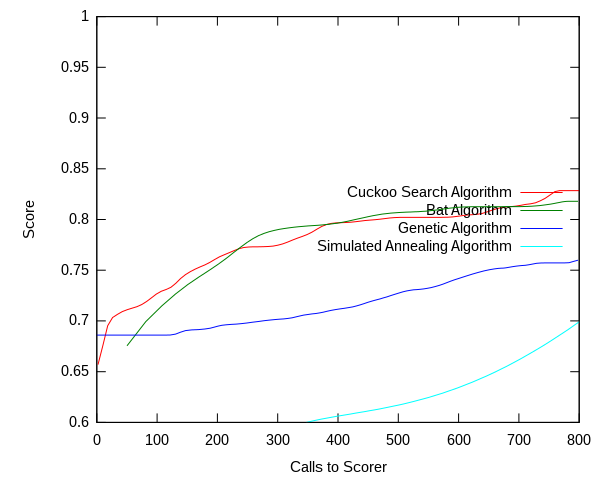}
\caption{Document 2}
\end{subfigure}
\begin{subfigure}[b]{0.38\textwidth}
\includegraphics[width=\textwidth]{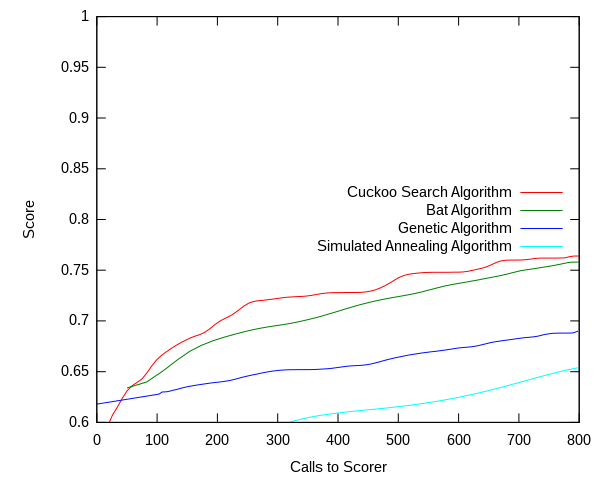}
\caption{Document 3}
\end{subfigure}
\begin{subfigure}[b]{0.38\textwidth}
\includegraphics[width=\textwidth]{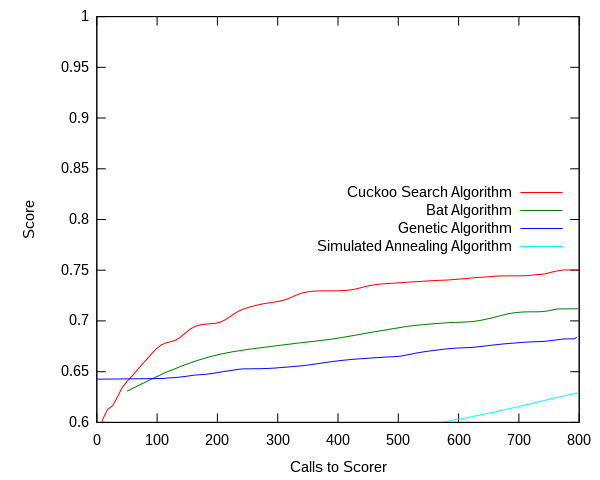}
\caption{Document 4}
\end{subfigure}
\begin{subfigure}[b]{0.38\textwidth}
\includegraphics[width=\textwidth]{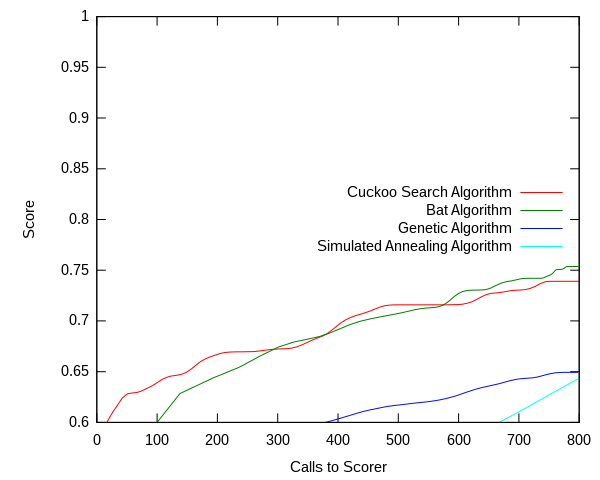}
\caption{Document 5}
\end{subfigure}
\end{center}
\caption{Evaluation of the global algorithms for 800 scorer calls}
\label{fig:ressco800}
\end{figure}

\newpage

\subsection{Number of call to the scorer : 2000}
\begin{figure}[H]
\begin{center}
\begin{subfigure}[b]{0.38\textwidth}
\includegraphics[width=\textwidth]{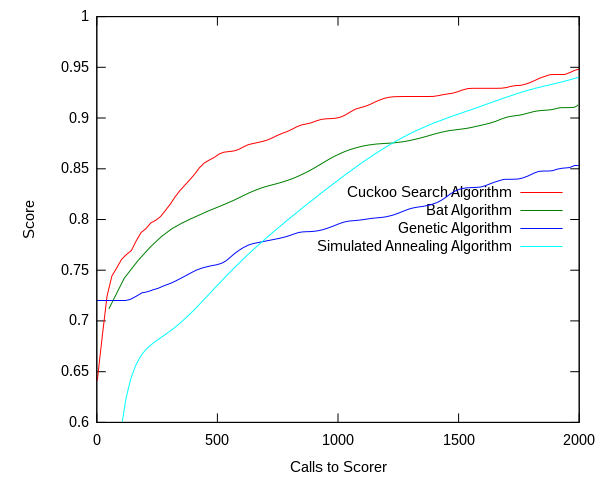}
\caption{Document 1}
\end{subfigure}
\begin{subfigure}[b]{0.38\textwidth}
\includegraphics[width=\textwidth]{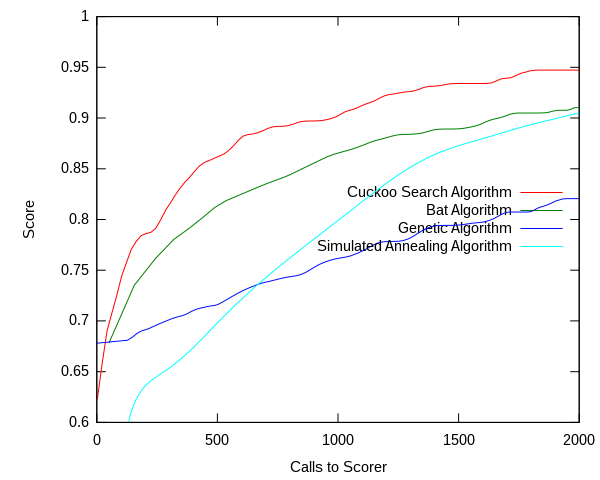}
\caption{Document 2}
\end{subfigure}
\begin{subfigure}[b]{0.38\textwidth}
\includegraphics[width=\textwidth]{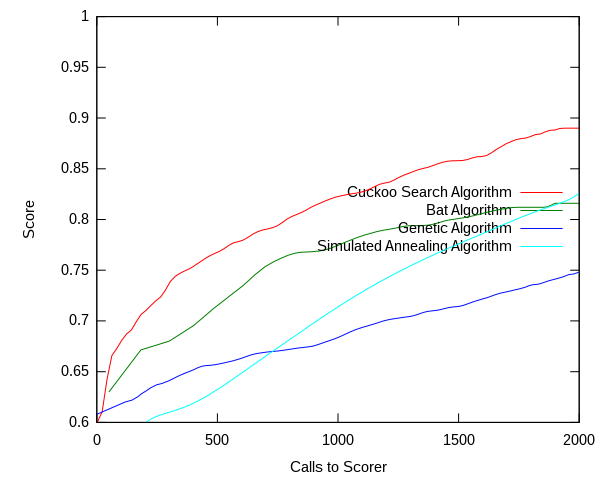}
\caption{Document 3}
\end{subfigure}
\begin{subfigure}[b]{0.38\textwidth}
\includegraphics[width=\textwidth]{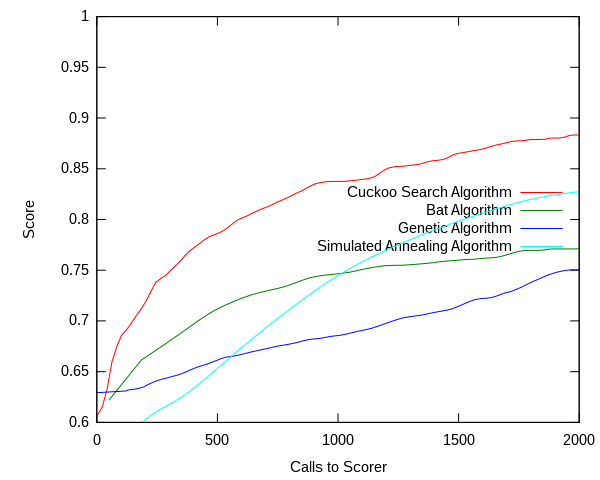}
\caption{Document 4}
\end{subfigure}
\begin{subfigure}[b]{0.38\textwidth}
\includegraphics[width=\textwidth]{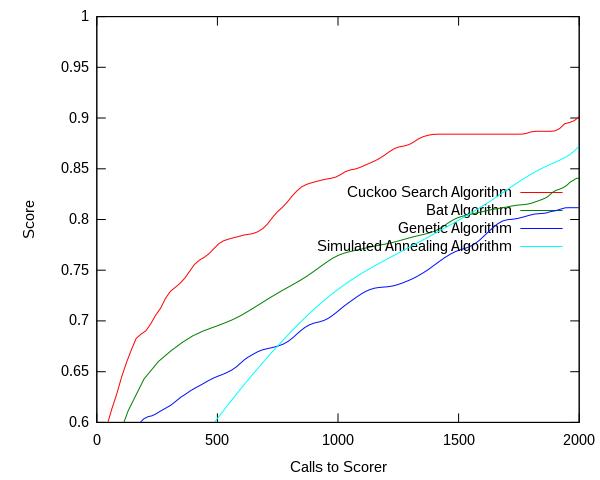}
\caption{Document 5}
\end{subfigure}
\end{center}
\caption{Evaluation of the global algorithms for 2000 scorer calls}
\label{fig:ressco2000}
\end{figure}

\newpage

\begin{figure}[H]
\begin{center}
\begin{subfigure}[b]{0.38\textwidth}
\includegraphics[width=\textwidth]{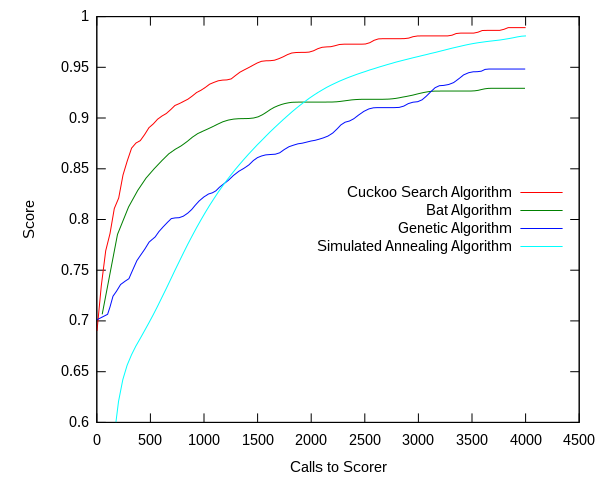}
\caption{Document 1}
\end{subfigure}
\begin{subfigure}[b]{0.38\textwidth}
\includegraphics[width=\textwidth]{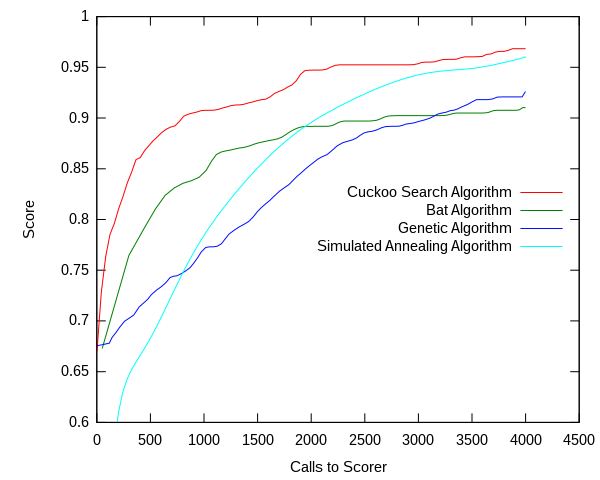}
\caption{Document 2}
\end{subfigure}
\begin{subfigure}[b]{0.38\textwidth}
\includegraphics[width=\textwidth]{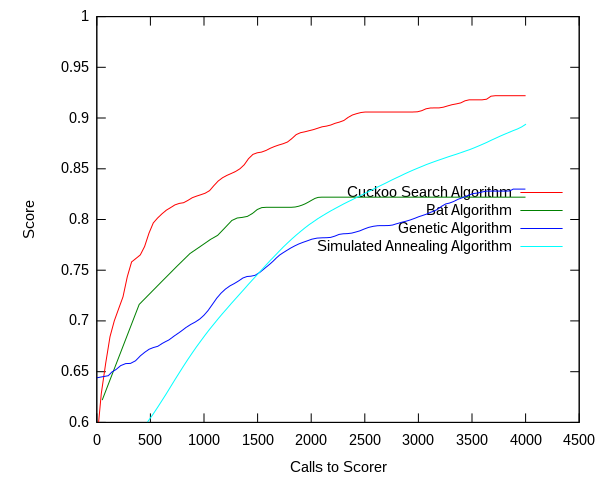}
\caption{Document 3}
\end{subfigure}
\begin{subfigure}[b]{0.38\textwidth}
\includegraphics[width=\textwidth]{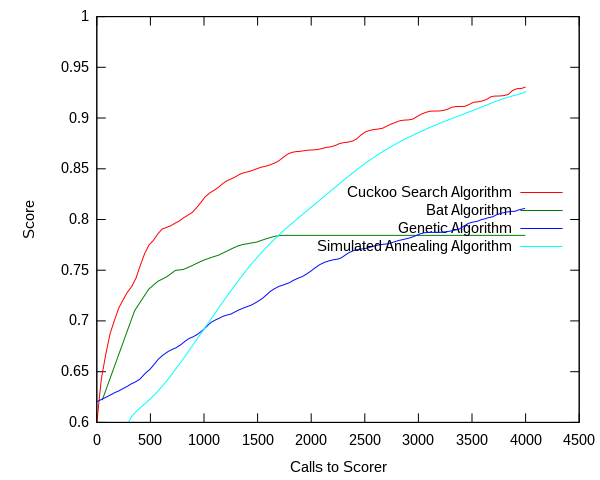}
\caption{Document 4}
\end{subfigure}
\begin{subfigure}[b]{0.38\textwidth}
\includegraphics[width=\textwidth]{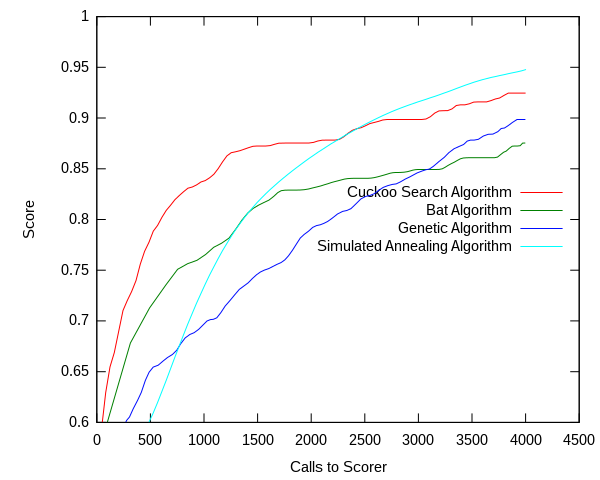}
\caption{Document 5}
\end{subfigure}
\end{center}
\caption{Evaluation of the global algorithms for 4000 scorer calls}
\label{fig:ressco4000}
\end{figure}

\newpage

\section{Implementation details}
\label{appendix:implem}
\subsection{The Bat Algorithm}
\lstset{
 basicstyle=\scriptsize,
 }

\begin{lstlisting}[language=java]
Configuration disambiguate(Document document)
{
    for (int i = 0 ; i < batsNumber ; ++i)
    {
         bats[i] = new Bat();
    }
    updateBestBat();
    
    while (!stopCondition.stop() && nbBatsFinished < batsNumber)
    {
        for (Bat currentBat : bats)
        {
            Configuration previousPosition = currentBat.position.clone();
            int previousVelocity = currentBat.velocity;
            double previousScore = currentBat.score;

            if (currentBat.rate < randomDoubleInRange(minRate, maxRate))
            {
                currentBat.position = bestBat.position.clone();
                currentBat.position.makeRandomChanges((int) getAverageLoudness());
            }
            else
            {
                currentBat.frequency = randomDoubleInRange(minFrequency, maxFrequency);
                for (int i = 0 ; i < dimension ; i++)
                {
                    if (currentBat.position.getAssignment(i) != bestBat.position.getAssignment(i))
                    {
                        currentBat.velocity++;
                    }
                }
                currentBat.velocity *= currentBat.frequency;
                currentBat.position.makeRandomChanges(currentBat.velocity);
            }

            if (currentBat.loudness >= randomDoubleInRange(minLoudness, maxLoudness) && 
            currentBat.score > bestBat.score)
            {
                currentBat.loudness *= alpha;
                if (currentBat.loudness < minLoudness) nbBatsFinished++;
                currentBat.rate = currentBat.initialRate * (1 - Math.exp(-gamma * currentIteration));
                bestBat = currentBat;
            }
            else
            {
                currentBat.position = previousPosition;
                currentBat.velocity = previousVelocity;
                currentBat.score = previousScore;
            }
        }
    }
    return bestBat.position;
}
\end{lstlisting}

\cleardoublepage

\subsection{The Cuckoo Search Algorithm}
\begin{lstlisting}[language=java]
Configuration run (Document document)
{    
    for (int i = 0 ; i < nestsNumber ; i++)
    {
        nests[i] = new Nest();
    }

    while (!stopCondition.stop())
    {
        int i = random.nextInt(nests.length);
        Nest new_i = nests[i].clone();
        new_i.randomFly();
            
        int j = random.nextInt(nests.length);
        while (j == i) j = random.nextInt(nests.length);
            
        if (new_i.getScore() > nests[j].getScore())
        {
            nests[j] = new_i;
        }
            
        sortNests();
        abandonWorthlessNests();
    }
    sortNests();
    return nests[nestsNumber - 1].configuration;
}

void abandonWorthlessNests()
{
    for (int i = 0 ; i < destroyedNestsNumber ; i++)
    {
        nests[i] = new Nest();
    }
}

void randomFly()
{
    double distance = levyDistribution.sample();
    configuration.makeRandomChanges((int) distance);
    needRecomputeScore = true;
}
        
public double getScore()
{
    if (needRecomputeScore)
    {
        score = configurationScorer.computeScore(currentDocument, configuration);
        needRecomputeScore = false;
    }
    return score;
}
\end{lstlisting}

\end{document}